\documentclass[runningheads]{llncs}

\usepackage{pdfpages}
\usepackage{cite}
\usepackage{amsmath,amssymb,amsfonts}
\usepackage{algorithmic}
\usepackage{graphicx}
\usepackage{textcomp}
\usepackage{xcolor}
\usepackage{soul}
\usepackage{url}
\usepackage[hidelinks]{hyperref}
\usepackage[utf8]{inputenc}
\usepackage{caption}
\usepackage{graphicx}
\usepackage{amsmath}
\usepackage{booktabs}
\usepackage{algorithm}
\usepackage{algorithmic}
\usepackage{subcaption}

\usepackage{longtable}
\usepackage{lscape}
\usepackage{booktabs}
\usepackage{tabularx}
\newcolumntype{Y}{>{\raggedleft\arraybackslash}X}
\usepackage{dcolumn}
\usepackage{supertabular}

\usepackage{amssymb}
\usepackage{amsmath}
\newcommand{\eg}{\textit{e.g.}}
\newcommand{\ie}{\textit{i.e.}}
\usepackage[switch]{lineno}

\newcommand{\qvec}{{\mathbf{q}}}
\newcommand{\xmat}{{\mathbf{X}}}
\DeclareMathOperator*{\argmin}{arg\,min}

\begin{document}
\title{Deep Depression Prediction on Longitudinal Data via Joint Anomaly Ranking and Classification}
\titlerunning{Deep Depression Prediction on Longitudinal Data}
%
\author{Guansong Pang\textsuperscript{\rm 1}, 
Ngoc Thien Anh Pham\textsuperscript{\rm 2}, 
Emma Baker\textsuperscript{\rm 2}, \\
Rebecca Bentley\textsuperscript{\rm 3}, 
Anton van den Hengel\textsuperscript{\rm 2} \\
}
\institute{Singapore Management University, Singapore\\ \email{gspang@smu.edu.sg} \and 
University of Adelaide, Australia \\
\email{\{ngoc.t.pham;emma.baker;anton.vandenhengel\}@adelaide.edu.au}
\and University of Melbourne, Australia\\
\email{brj@unimelb.edu.au}
}
\authorrunning{G. Pang et al.}
%
%
\maketitle              
\begin{abstract}
A wide variety of methods have been developed for identifying depression, but they focus primarily on measuring the degree to which individuals are suffering from depression currently. In this work we explore the possibility of predicting future depression using machine learning applied to longitudinal socio-demographic data. In doing so we show that data such as housing status, and the details of the family environment, can provide cues for predicting future psychiatric disorders. To this end, we introduce a novel deep multi-task recurrent neural network to learn time-dependent depression cues.  The depression prediction task is jointly optimized with two auxiliary anomaly ranking tasks, including contrastive one-class feature ranking and deviation ranking. The auxiliary tasks address two key challenges of the problem: 1) \textit{the high within class variance of depression samples}: they enable the learning of representations that are robust to highly variant in-class distribution of the depression samples; and 2) \textit{the small labeled data volume}: they significantly enhance the sample efficiency of the prediction model, which reduces the reliance on large depression-labeled datasets that are difficult to collect in practice. Extensive empirical results on large-scale child depression data show that our model is sample-efficient and can accurately predict depression 2-4 years before the illness occurs, substantially outperforming eight representative comparators.

\end{abstract}

\section{Introduction}

Major Depressive Disorder (MDD), widely known as depression, is a mental disorder characterized by a severe and persistent feeling of sadness, loss of interest in activities, or a sense of despair, causing significant impairment in daily life \cite{lamers2019longitudinal}. Globally over 300 million people of all ages are estimated to suffer from depression, and it is a major contributor to nearly 800 thousands suicide deaths per year \cite{world2017depression}. There have been many studies \cite{nasir2016multimodal,gong2017topic,shen2018cross,uddin2020depression} demonstrating effective automated diagnosis of depression using machine learning techniques. These studies  focus on the detection of ongoing, current depression, using data that relates information about the current state of an individual.  We focus here on the more challenging task of predicting depression in advance.  This is achieved using longitudinal socio-demographic data. 
The motivation for this approach is that prediction far enough in advance of upcoming depression might enable early intervention before the condition arises.

The longitudinal socio-demographic data used contains a variety of non-medical information such as education level, socio-economic background, family environment, measured at intervals overtime.  There are three  major challenges in exploiting these cues for depression prediction. First, this is noisy high-dimensional temporal data that contains thousands of numeric and categorical features, among which only very selective set are relevant to depression prediction. Second, there is no single set of socio-demographic factors that cause depression, leading to high in-class variance for the depression samples. For example, for childhood depression, the root cause may be related to the mental health status of parents, dwelling conditions, or a medical condition.  As a result, the depression cases are highly dissimilar and exhibit no single underlying mechanism, or common characteristics. A model that seeks a single common explanation for all cases cannot succeed.
Lastly, in practice, only a small amount of labeled data is available, as it is difficult, if not impossible, to collect large depression samples. The available  data thus often fails to cover the diverse types of depression cases.

In this work we introduce a novel multi-task learning approach to tackle these challenges, in which a depression classification task is jointly optimized with two auxiliary anomaly ranking tasks, contrastive one-class feature ranking and deviation ranking. Depression samples are treated as anomalous samples in our auxiliary tasks, because the cause-varying depression samples can be widely distributed, and as a result,  are difficult to  model as a single concrete class. Directly modeling these depression samples with classification models can easily overfit the given depression cases. The two anomaly ranking tasks are devised to enforce compact low-dimensional representations of normal samples and allow variations in the representations of depression samples, presenting effective inductive biases to regularize the classification models. This significantly improves the model's generalization beyond the individual cases presented in the small volume of labeled data available.

In summary, this work makes two key contributions:
\begin{itemize}
    \item We introduce a novel multi-task learning framework, which harnesses auxiliary anomaly detection tasks to empower the greater classification task. To the best of our knowledge, this is the first multi-task learning approach that jointly optimizes classification and anomaly ranking tasks, which is an important tool for application problems that are similar to depression prediction.
    \item We further instantiate the framework into a multi-task recurrent neural network model, termed MTNet, which optimizes the depression model with one-class constraints on its feature space and deviation constraints on its output layer. Extensive empirical results on large-scale child depression data show that MTNet can accurately predict depression two to four years before the illness occurs (\eg, achieving a recall of 0.8), substantially outperforming eight competing methods. Additionally, MTNet can also outperform these competing methods even with largely reduced (50\% less) training data.
\end{itemize}

\section{Related Work}

\noindent \textbf{Longitudinal Studies} There have been many longitudinal studies of depression \cite{henry2018screening,korsten2019factors}, but they focus on association discovery that identifies factors or predictors associated with depression using traditional bivariate/multivariate statistic models. By contrast, our study is on learning prediction models to predict the occurrence/risk of future depression. 

\vspace{0.1cm}
\noindent \textbf{Automated Depression Diagnosis} 
Current studies focus on the detection of depression using classification models on vocal/visual data taken during clinical interviews, with vocal features like prosodic and cepstral features \cite{yang2012detecting,nasir2016multimodal,gong2017topic} and visual features like facial expression, gaze direction, and eye movement \cite{nasir2016multimodal,zhou2018visually,uddin2020depression}. Depression detection based on social media data using linguistic and network features is also extensively studied \cite{shen2018cross,mann2020see}. However, all these studies focus on the detection of ongoing depression, while we aim at predicting upcoming depression.

An exploratory task is introduced at the eRisk workshop \cite{losada2018overview} to facilitate the development and evaluation of models for early detection of signs of depression using social media data, resulting in a number of early depression detection models \cite{sadeque2018measuring,masood2019adapting,burdisso2020tau}. Like the aforementioned studies, they are also focused upon detecting \textit{expression of depression cues/symptoms}, while we learn the \textit{hidden root cause factors in socio-demographic features} which are more difficult to learn.

\vspace{0.1cm}
\noindent \textbf{Multi-task Learning} Multi-task learning \cite{ruder2017overview} has been successful in a range of applications. There have been many approaches introduced to jointly learn multiple related tasks (e.g., classification, regression, and clustering) to improve the performance on small labeled data, such as feature learning, task relation learning, task clustering, low-rank and decomposition approaches \cite{zhang2017survey}. Our work uses multi-task feature learning methods with hard parameter sharing as in \cite{strezoski2019many,hassani2019unsupervised,standley2020tasks}, but our approach is the first work that uses anomaly ranking at the feature and output layers to regularize the classification.

\vspace{0.1cm}
\noindent \textbf{Anomaly Detection} Anomaly detection techniques have been successfully applied to detect abnormal events/behaviors in many applications \cite{pang2021deep}, but they are rarely used to regularize supervised learning models in multi-task learning as the two paradigms have rather different learning objectives. We show in this work that recent advanced anomaly detection models \cite{ruff2018deep,pang2019deep} can be adapted to effectively regularize classification models and improve their sample efficiency.

\section{Multi-task Recurrent Neural Networks}

\subsection{The Proposed Framework}

Upcoming depression prediction aims to learn a binary depression classification mapping function $\phi:\mathcal{X}\rightarrow \mathcal{Y}$, where $\mathcal{X}=\{\mathbf{X}_{1}, \mathbf{X}_{2},  \cdots,\mathbf{X}_{N}\}$ is a set of longitudinal data of $N$ samples; each sample $\mathbf{X}\in\mathbb{R}^{w \times D}$ is a matrix input of an individual subject, which contains the socio-demographic features of the subject in the recent $w$ \textit{waves} (or time steps) of questionnaire data, \ie, $\mathbf{X}=\{\mathbf{x}_1, \mathbf{x}_2,\cdots, \mathbf{x}_w\}$, where $\mathbf{x}_t\in \mathbb{R}^{D}$ is a feature vector derived from the $t$-th wave of questionnaire data; $\mathcal{Y}=\{0,1\}$ is the output space, with `1' indicating the subject being normal within all the recent $w$ waves of questionnaire but having depression in the next waves of questionnaire, and with `0' indicating the subject being normal in the recent and future waves of questionnaire.

\begin{figure}[h!]
  \centering
    \includegraphics[width=0.75\textwidth]{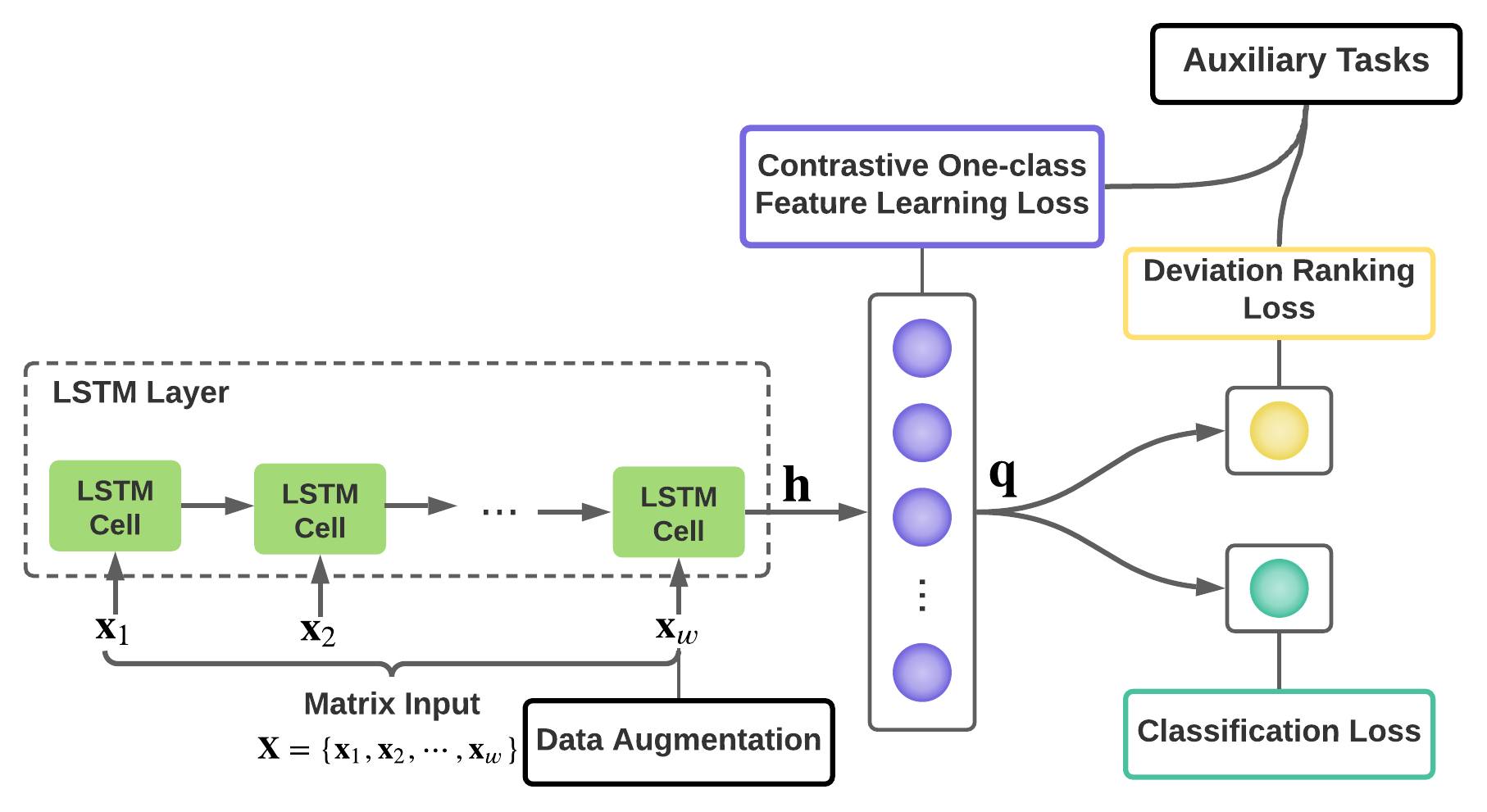}
  \caption{The Proposed Multi-task Learning Framework. }
  \label{fig:mtnet}
\end{figure}

In this work we introduce a novel multi-task learning framework to tackle the problem. An overview of the approach is presented in Figure \ref{fig:mtnet}. Depression classification is our primary task and is jointly optimized with two auxiliary tasks, including deviation score ranking and contrastive one-class feature learning. The auxiliary tasks treat depression samples as anomalies and enforce compact feature representations of normal samples and allow some variations in the representations of depression samples, serving as a regularizer of the classification model. This results in better generalized classification models than that in the single primary task. Formally, let $\tau: \mathcal{X} \rightarrow \mathbb{R}$ be an anomaly ranking function that assigns an anomaly score to each subject; $\psi: \mathcal{X} \rightarrow \mathcal{Q}$ be the one-class feature learning function, where $\mathcal{Q} \in \mathbb{R}^{M}$ with $M \ll D$ is a new feature space, then our overall objective function can be given as follows.
\begin{multline}
    \argmin_{\Theta_{e}, \Theta_a, \Theta_o} \sum_{i=1}^{N} \bigg[ \ell_{e}\big(\phi(\mathbf{X}_i; \Theta_{e}), y_i\big) + \alpha \ell_{a}\big(\tau(\mathbf{X}_i; \Theta_a), y_i \big) + \beta \ell_{o}\big(\psi(\mathbf{X}_i; \Theta_o) , y_i\big) \bigg],
\end{multline}
where $\ell_{e}$, $\ell_{a}$ and $\ell_{o}$ are respective loss functions for depression classification, deviation ranking and contrastive one-class metric learning, $y_i$ is the class label of $\xmat_i$, $\Theta=\{\Theta_{e}, \Theta_a, \Theta_o\}$ is the set of parameters to be learned, $\alpha$ and $\beta$ are hyperparameters to control the importance of the two auxiliary tasks.

We instantiate the framework into a model called MTNet that leverages a shared LSTM neural network \cite{hochreiter1997long} to learn critical temporal changes in longitudinal data for depression classification. Supervised anomaly deviation and one-class support vector data description loss functions are defined to improve the model's generalization. A simple data augmentation method is also introduced to further enhance the generalizability. The modules of MTNet are presented as follows.

\subsection{Primary Task: Depression Classification}

Our classification model leverages an LSTM neural network layer to learn important temporal dependency in the longitudinal data. An LSTM layer consists of multiple LSTM cells, with each LSTM cell learning temporal-dependent representations of the input data at a specific wave. The full LSTM layer uses the recurrent LSTM cells to encode important temporal changes across all different questionnaire waves into the output vector $\mathbf{h} \in \mathbb{R}^{L}$ in the last LSTM cell, \ie, the LSTM layer is a mapping function $\eta$ that performs $\mathbf{h}=\eta(\mathbf{X};\Theta_l)$, where $\Theta_l$ contains all weight parameters in a standard LSTM (see \textit{Supplementary Material}\footnote{Supplementary material is available at \url{https://tinyurl.com/MTNetPAKDD22}} for the full details of LSTM). To learn more expressive representations, a FC layer is used to project $\mathbf{h}$ onto a lower-dimensional feature representation space:
\begin{equation}\label{eq:feat}
    \qvec=\psi(\xmat;\Theta_l,\mathbf{W}_{s},\mathbf{b}_s)= g_{s}(\mathbf{W}_{s}\eta(\mathbf{X};\Theta_l) + \mathbf{b}_s),
\end{equation}
where $\mathbf{W}_{s}  \in \mathbb{R}^{M\times L}$ and $\mathbf{b}_{s}\in \mathbb{R}^{M}$ are the learnable parameters, $g_{\text{s}}$ is an activation function, and $\qvec\in \mathcal{Q}$ is the final feature representation of $\mathbf{X}$.
We then train a classifier on the $\mathcal{Q}$ representation space with a standard binary cross-entropy loss function:
\begin{equation}\label{eqn:ce}
    \ell_{e}(\xmat, y) = -\big( y\log(p) + (1-y)\log(1-p) \big),
\end{equation}
where $y$ is the class label of $\xmat$ and $p=\phi(\xmat; \Theta_{e}) = g_{e}(\mathbf{W}_{e}\mathbf{q} + b_e)$, where $g_{e}$ is a sigmoid activation, $\mathbf{W}_{e}  \in \mathbb{R}^{1\times M}$ and $b_{e}\in \mathbb{R}$. $\Theta_{e}=\{\Theta_l,\mathbf{W}_{s}, b_{s},\mathbf{W}_{e}, b_{e}\}$ contains all the network parameters that can be learned in an end-to-end manner. 

\subsection{Auxiliary Tasks} 
Two auxiliary tasks, contrastive one-class-based feature ranking and deviation ranking, are incorporated to introduce an inductive bias (\ie, to learn compact normal representations and allow large variations in abnormal representations) to learn more generalized representations of depression samples. 

\vspace{0.1cm}
\noindent\textbf{Deviation Ranking} A partially-supervised anomaly ranking task is introduced to enforce the model to assign significantly larger anomaly scores for depression samples than that of non-depression samples. Inspired by \cite{pang2019deep}, a prior-driven anomaly ranking loss function, called deviation loss, is used to fulfill this goal. Particularly, a Gaussian prior $\mathcal{N}(\mu, \sigma^2)$ is imposed on the anomaly scores of all samples, which posits that the anomaly scores of non-depression samples are centered around a Gaussian mean value $\mu$ while the anomaly scores of depression samples have at least $a * \sigma$ deviations from $\mu$. 
Formally, we add another network output head with one linear unit to learn an anomaly score for each sample:
\begin{equation}
    \tau(\xmat; \Theta_a) = \mathbf{W}_a\psi(\xmat;\Theta_r)+b_a,
\end{equation}
where $\Theta_a=\{\Theta_r, \mathbf{W}_a,b_a\}$ are the parameters to be learned. We then define the deviation using the well-known Z-Score:
\begin{equation}
    \mathit{dev}(\xmat) = \frac{ \tau(\xmat; \Theta_a)- \mu}{\sigma},
\end{equation}
The deviation function is then used to define our anomaly ranking loss function:
\begin{equation}\label{eqn:loss}
    \ell_a(\xmat, y) = (1-y)|\mathit{dev}(\xmat)| + y \max\big(0, a - \mathit{dev}(\xmat)\big).
\end{equation}
By minimizing $\ell_a$, our model pushes the anomaly scores of normal samples as close as possible to $\mu$ while enforcing at least $a * \sigma$ between $\mu$ and the anomaly scores of depression samples in the upper tail of the Gaussian distribution. Following \cite{pang2019deep}, the prior $\mathcal{N}(0, 1)$ is used with $a=5$ to guarantee significant deviations of depression samples from normal samples.


\vspace{0.1cm}
\noindent\textbf{Contrastive One-class Feature Learning} Unlike the anomaly ranking task that introduces the inductive bias using an output layer independent from the classification output, the one-class feature learning complements the deviation score learning and exerts directly on the feature layer. Particularly, we introduce a contrastive one-class feature learning method, in which we devise a supervised variant of support vector data description (SVDD) \cite{ruff2018deep} by contrasting the one-class center of the normal samples and the depression samples.
\begin{multline}
    \ell_o(\xmat, y)= (1-y)||\psi(\xmat;\Theta_r) - \mathbf{n}||_{2} + y \max\big(0, m - ||\psi(\xmat;\Theta_r) - \mathbf{n}||_{2}\big),
\end{multline}
where $\Theta_o=\{\Theta_r\}$, $\mathbf{n}\in \mathbb{R}^{M}$ is the one-class center vector of normal samples and $m$ is a hyperparameter to control the contrast margin. The first term is the original SVDD objective. The second term is added to enforce a large margin between non-depression and depression samples in the $\psi$-induced representation space, while minimizing the $\mathbf{n}$-centered hypersphere’s volume. We found empirically that MTNet can perform well with varying settings of $\mathbf{n}$, \eg, $\mathbf{n} \sim \mathcal{N}(0,1)$ or $\mathbf{n} \sim \mathcal{U}(0,1)$. We use $\mathbf{n} \sim \mathcal{N}(0,1)$ by default, \ie, generating $\mathbf{n}$ by randomly drawing a vector from a standard Gaussian distribution. $m=1$ is used to enforce a sufficiently large distance margin in the feature representation space.

\subsection{Data Augmentation} A simple data augmentation method is introduced to augment depression samples and further enhance the model's generalizability. Specifically, a pair of depression samples are randomly selected, and then a small percentage of randomly selected values in the last wave data of one sample are replaced with the corresponding values in another sample to create a new depression sample. The augmented sample can well retain the original depression-relevant information while at the same time enriching the depression samples. By using this method, we increase the number of depression samples in the training data by a factor of 10. In our experiment, we randomly replaced 5\% of the feature values by default. 

\subsection{The Algorithm of MTNet}

The algorithmic procedure of our model MTNet is presented in Algorithm \ref{alg:mtnet}. After random initialization of the network parameters in Step 1, stochastic gradient descent is used to optimize the model in Steps 2-8. In Step 4, as the number of depression samples is typically far smaller than that of non-depression samples, we generate sample batches with balanced class distribution to achieve more effective optimization. This shares the same spirit as oversampling in imbalanced learning \cite{he2009learning}. Step 5 calculates the batch-wise loss for the three tasks. Step 6 performs gradient descent steps to learn the parameters $\Theta$. Note that $\Theta_r$ are shared parameters in $\{\Theta_{e}, \Theta_a, \Theta_o\}$, and thus, the feature representations in MTNet are jointly optimized by all three tasks. At the inference stage, only the classification function $\phi$ is used to produce the class label.

\vspace{-0.5cm}
\renewcommand{\algorithmicrequire}{\textbf{Input:}}
\renewcommand{\algorithmicensure}{\textbf{Output:}}
\begin{algorithm}
\small 
\caption{\textbf{MTNet}}
\begin{algorithmic}[1]
\label{alg:mtnet}
\REQUIRE $\mathcal{X} \in \mathbb{R}^{w\times D}$ - training samples, and binary class labels $\mathbf{y}$
\ENSURE $\phi: \mathcal{X} \rightarrow \mathcal{Y}$ - a depression classification network
\STATE Randomly initialize $\Theta=\{\Theta_{e}, \Theta_a, \Theta_o\}$
\FOR{ $j = 1$ to $\mathit{\# epochs}$}
    \FOR{ $k = 1$ to $\mathit{\# batches}$}
        \STATE $\mathcal{B} \leftarrow$ Randomly sample the same number of depression and non-depression samples
        \STATE Calculate the loss using $\frac{1}{|\mathcal{B}|}\sum\limits_{\xmat_i \in \mathcal{B}} \bigg[ \ell_{e}\big(\phi(\mathbf{X}_i; \Theta_{e}), y_i\big) + \alpha \ell_{a}\big(\tau(\mathbf{X}_i; \Theta_a), y_i \big) + \beta \ell_{o}\big(\psi(\mathbf{X}_i; \Theta_o) , y_i\big) \bigg]$    
        \STATE Perform a gradient descent step w.r.t. the parameters in $\Theta$
    \ENDFOR
\ENDFOR
\RETURN $\phi$
\end{algorithmic}
\end{algorithm}
\vspace{-0.5cm}

\section{Experiments}

\subsection{Datasets}
Our model is evaluated on a large child depression data dataset based on the Longitudinal Study of Australian Children (LSAC) data \cite{gray2005growing}. LSAC consists of multiple bi-annual waves of questionnaire-based interview data of 10,090 children across Australia. Initially, children aged from infant to 5 years and their families are interviewed between August 2003 and February 2004. This routine is repeated every two years afterwards. At the time of writing seven waves of data are available. LSAC provides a dataset of 4,983 children aged 4 to 5 years in the first wave of questionnaire. These children are all healthy until 287 children are confirmed to have depression at the 6/7-th wave of interview in the years 2013 to 2015. Depression is measured using parental self-report data. Particularly, in the waves 6 and 7, the primary caregiver is asked ``does study child have any of these ongoing (depression) conditions\footnote{`Ongoing conditions' means that the conditions ``\textit{exist for some period of time (weeks, months or years) or re-occur regularly. They do not have to be diagnosed by a doctor}''.}?''; the child is confirmed to have depression if the answer is `yes'. After a simple feature screening to remove uninformative features (\eg, features with very large percentage of missing values), 210 social-demographic features related to individual growth and development (\eg, age, gender, living location, schooling performance), and family environment (\eg, social, educational, economic, employment, household income, housing conditions) are used (see \textit{Supplementary Material} for the list of the interview questions and a sample of the questionnaire). In the selected features, missing values are filled with the mean/mode value in each feature; categorical features are then converted into numeric features by using one-hot encoding. The resulting dataset contains 762 features in each wave of data. Thus, the dataset used has 4,983 samples, with each sample represented by a $7\times 762$ matrix. We further perform a stratified random split of the dataset into three subsets, including 60\% data as a training set, and respective 20\% data for validation and testing sets.

\subsection{Experimental Setup}
We evaluate the performance of predicting the possible occurrence of depression in the near future. To this end, the model is trained and tested using only the first five waves of data when all children are reported to be mentally healthy (\ie, absence of depressive symptoms). 
The task is to predict whether a child will have depression at the upcoming wave 6 or 7. MTNet is compared with eight temporal and non-temporal methods.

\begin{itemize}
    \item \textbf{Non-temporal Methods}. Three popular classification methods, including logistic regression (LR), support vector machines (SVM) and multi-layer perceptron (MLP) neural networks, are used as baseline methods that are not designed to capture temporal dependence. They are two main ways to apply these methods to the longitudinal data. One way is to build the classification model using the most recent single (\ie, 5-th) wave data only. The second way is to use the data from all the five waves, in which for each subject we concatenate the feature vectors derived from all the waves into one lengthy unified feature vector; the classifiers are then built upon this concatenated data. This way helps capture some temporal-dependent changes. All three methods are evaluated in both ways, with LR/SVM/MLP-s denoting the classifier using the single wave data and LR/SVM/MLP-m denoting the use of the concatenated multi-wave data.
    \item \textbf{Temporal Methods}. The LSTM-based deep classifier, is used as a competing temporal method. The standard binary cross-entropy loss function
    is used to train the model. We also compare MTNet with the state-of-the-art anomaly detection model DevNet \cite{pang2019deep} that is adapted to temporal data with LSTM network.
\end{itemize}

\subsection{Implementation Details}

MTNet is implemented with one LSTM layer with 200 units, followed by a fully-connected (FC) layer with 20 units and a classification output layer. The sigmoid function is used in $g_r$ in the LSTM layer by default; the widely-used ReLU activation function is used in $g_s$ in the FC layer. A dropout layer with a dropout rate of 0.5 is applied to the LSTM and FC layers. 
The competing methods LSTM and DevNet use exactly the same network architecture as MTNet.
MLP uses a similar network structure with two hidden layers of respectively 200 and 20 units, with each layer having a dropout rate of 0.5. MTNet, DevNet, LSTM, and MLP are implemented using Keras\footnote{https://keras.io/} and optimized using RMSprop with a batch size of 256 and 20 batches per epoch. They are trained with 30 epochs as their performance can converge early. $\alpha=0.5$ and $\beta=2.0$ are used in MTNet by default. LR and SVM are taken from the open-source scikit-learn package\footnote{https://scikit-learn.org/}. Due to a large percentage of irrelevant features presented in the data, our extensive results showed that applying the $l_1$-norm regularizer to MLP, LR and SVM obtains significantly better performance than the $l_2$-norm regularizer. Thus, the $l_1$-norm regularizer is applied to these three classifiers to bring sparsity to the model. The regularization hyperparameter is probed with $\{0.001, 0.01, 0.1, 1\}$, with the best performance reported. The oversampling method in MTNet is used in all competing methods to alleviate the class-imbalanced problem.


\subsection{Performance Evaluation Measures}
Three widely-used evaluation measures are used, including the Area Under Receiver Operating Characteristic Curve (AUC-ROC), Area Under Precision-Recall Curve (AUC-PR), and F$_{1}$-score (F-score for brevity). AUC-ROC summarizes the ROC curve of true positives against false positives, while AUC-PR summarizes the curve of precision against recall. AUC-ROC is popular due to its good interpretability. AUC-PR is more indicative than AUC-ROC in evaluating performance on imbalanced data. F-score is the harmonic mean of precision and recall. We also report the precision and recall results to gain more insights into the performance. 
The reported results are averaged over five independent runs. 

\subsection{Empirical Results} 

\noindent \textbf{Effectiveness on Real-world Data} The results of depression prediction are shown in Table \ref{tab:prediction}. Our model MTNet is the best performer in AUC-ROC, AUC-PR and F-score. MTNet substantially outperforms all of its competing methods by 2\%-38\% in AUC-PR and 3\%-20\% in F-score. Impressively, MTNet obtains a recall of 0.8, achieving at least 7.8\% improvement over its contenders. Given a sufficiently high precision of 0.7, the high recall rate in MTNet would enable accurate intervention and pre-treatment of up to 80\% upcoming depression cases at a very early stage (2-4 years before the depression occurs), effectively preventing and reducing the depression cases.

\begin{table}[htbp]
\centering
\caption{Performance Results (mean$\pm$std) of Depression Prediction.}
\scalebox{0.95}{
\begin{tabular}{cccccc}
\hline
 & \centering \textbf{AUC-ROC} & \centering \textbf{AUC-PR} & \centering \textbf{F-score} & \centering \textbf{Precision} & \textbf{Recall} \\ \hline
\textbf{LR-s} & 0.648$\pm$0.020 & 0.595$\pm$0.018 & 0.632$\pm$0.021 & 0.659$\pm$0.032 & 0.611$\pm$0.037 \\ 
\textbf{LR-m} & 0.648$\pm$0.019 & 0.596$\pm$0.016 & 0.620$\pm$0.023 & 0.670$\pm$0.027 & 0.579$\pm$0.038 \\ 
\textbf{SVM-s} & 0.666$\pm$0.012 & 0.610$\pm$0.012 & 0.646$\pm$0.005 & 0.682$\pm$0.023 & 0.614$\pm$0.011 \\ 
\textbf{SVM-m} & 0.684$\pm$0.013 & 0.631$\pm$0.010 & 0.646$\pm$0.026 & 0.729$\pm$0.013 & 0.582$\pm$0.045 \\ 
\textbf{MLP-s} & 0.771$\pm$0.007 & 0.780$\pm$0.010 & 0.706$\pm$0.023 & 0.679$\pm$0.032 & 0.742$\pm$0.071 \\ 
\textbf{MLP-m} & 0.814$\pm$0.009 & 0.808$\pm$0.019 & 0.718$\pm$0.022 & 0.730$\pm$0.049 & 0.718$\pm$0.082 \\ 
\textbf{LSTM} & 0.779$\pm$0.012 & 0.775$\pm$0.010 & 0.662$\pm$0.034 & \textbf{0.751}$\pm$0.031 & 0.593$\pm$0.039 \\ 
\textbf{DevNet} & 0.785$\pm$0.013 & 0.786$\pm$0.017 & 0.688$\pm$0.035 & 0.704$\pm$0.044 & 0.684$\pm$0.094 \\ 
\textbf{MTNet} & \textbf{0.818}$\pm$0.009 & \textbf{0.823}$\pm$0.008 & \textbf{0.743}$\pm$0.037 & 0.697$\pm$0.006 & \textbf{0.800}$\pm$0.080\\ \hline
\end{tabular}
}
\label{tab:prediction}
\end{table}

\begin{figure}[t]
\centering
\begin{minipage}[b]{0.42\linewidth}
  \centering
  \centerline{\includegraphics[width=1.05\linewidth]{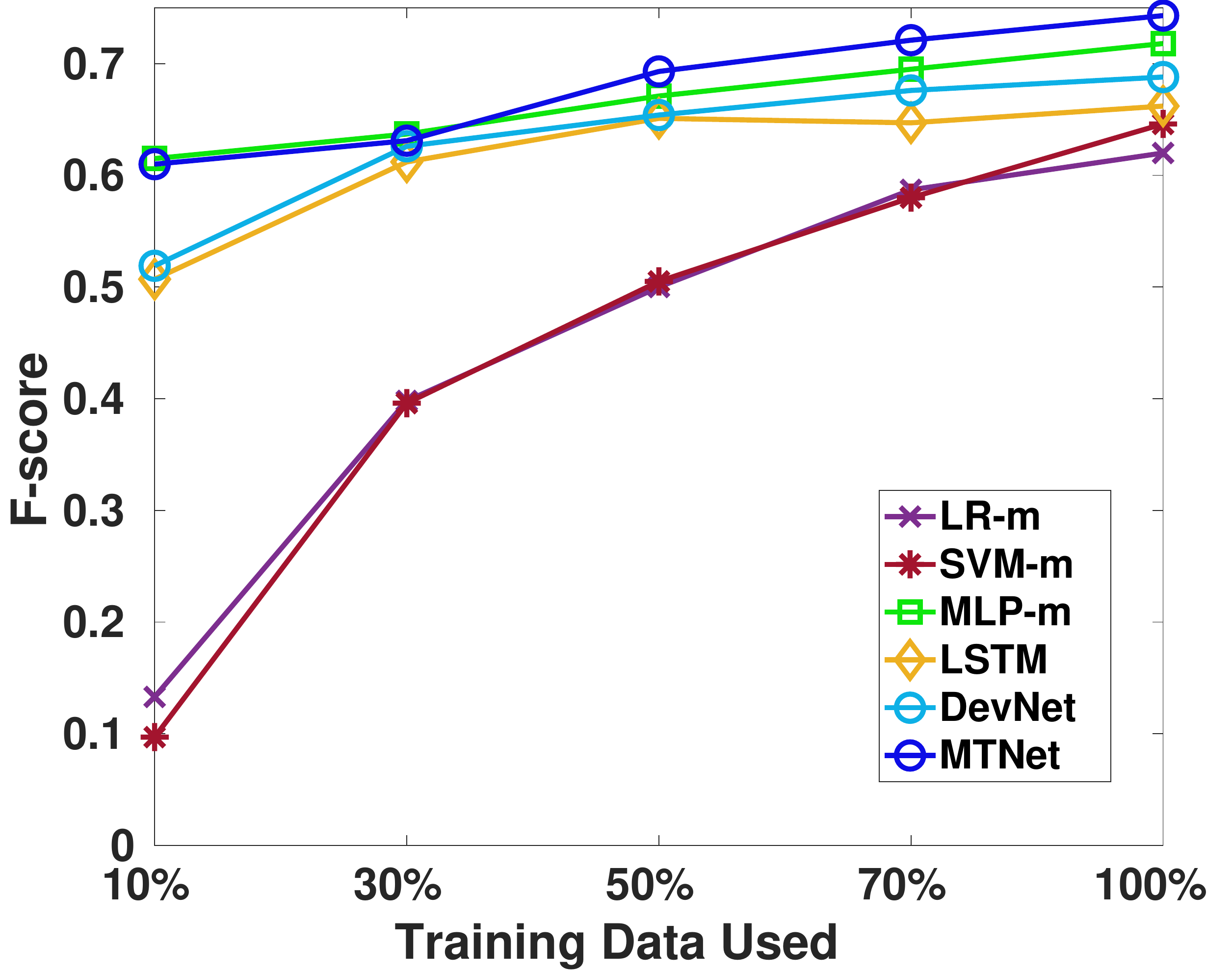}}
  \captionof{figure}{Sample Efficiency}\medskip
  \label{fig:sample_efficiency}
\end{minipage}
\hfill
\begin{minipage}[b]{0.42\linewidth}
  \centering
  \centerline{\includegraphics[width=1.05\linewidth]{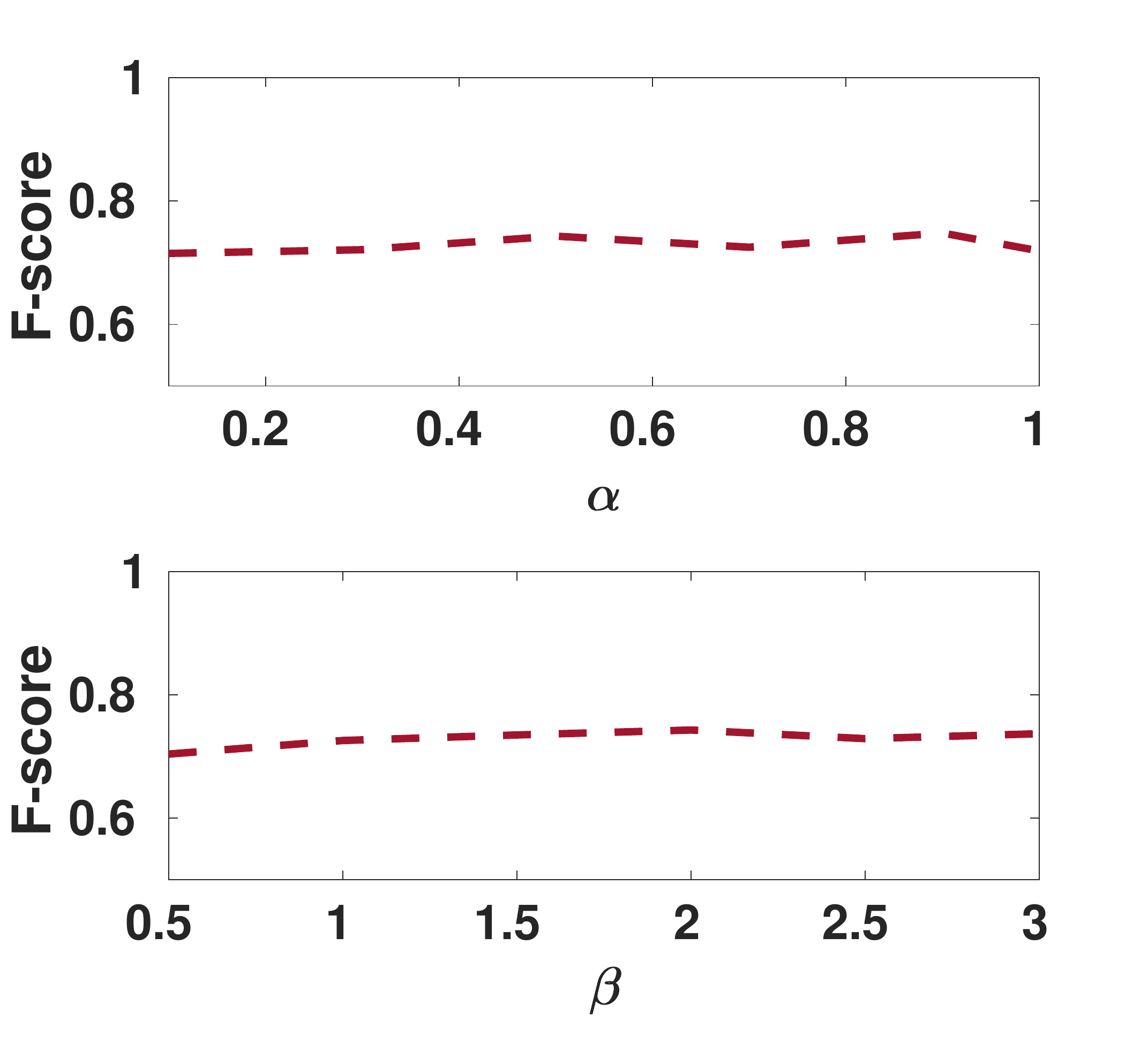}}
  \captionof{figure}{Parameter Sensitivity}\medskip
  \label{fig:sensitivity}
\end{minipage}

\end{figure}

\vspace{0.1cm}
\noindent \textbf{Sample Efficiency}
This section examines the model's generalizability from the sample efficiency aspect, \ie, how is the performance if less labeled training data is available? Specifically, each model is trained on a new training set that is a random subset of the original training data, and then it is evaluated using the same test data as that used in Table \ref{tab:prediction}. We focus on comparing to the better contenders LR/SVM/MLP-m and omit the less effective ones -- LR/SVM/MLP-s.
The results are shown in Figure \ref{fig:sample_efficiency}. Remarkably, MTNet is substantially more sample-efficient than its competing methods; it performs much better than, or comparably well to, the best performance of its competing methods even when it uses 50\% less training data. This superiority of MTNet benefits from the integrated anomaly ranking tasks in its multi-task objective function, which enable better generalization to diverse depression cases. This is manifested by the large performance gap between MTNet and LSTM, since the only difference between them is the two auxiliary tasks integrated into MTNet.  

\vspace{0.1cm}
\noindent \textbf{Parameter Sensitivity}
This section evaluates the sensitivity of MTNet w.r.t. its two hyperparameters, $\alpha$ and $\beta$, which respectively adjust the importance of the anomaly ranking loss and one-class metric loss. The test results are presented in Figure \ref{fig:sensitivity}. The results show that MTNet generally performs stably with both $\alpha$ and $\beta$ in a wide range of setting choices. Relatively small $\alpha$ and large $\beta$ are needed for MTNet to achieve the best performance. This indicates that MTNet is dependent more on the one-class feature learning than the deviation ranking. 

\vspace{0.1cm}
\noindent \textbf{Ablation Study}
This section evaluates the importance of each module in MTNet. LSTM is used as a baseline to evaluate the effect of incorporating one or more of the following modules, including the deviation ranking loss $\ell_a$, the one-class metric learning loss $\ell_o$, and the data augmentation (DA). The results are reported in Table \ref{tab:ablation}. It is clear that i) the multi-task learning performs substantially better than the individual tasks, ii) all three modules in MTNet make important contribution to its overall performance, and iii) the deviation ranking loss $\ell_a$ and the one-class metric learning loss $\ell_o$ are complementary to each other. 


\begin{table}[htbp]
\centering
\caption{Ablation Study Results. DA is short for data augmentation.}
\scalebox{0.95}{
\begin{tabular}{l@{}ccccc}
\hline\hline
Method & \textbf{AUC-ROC}  & \textbf{AUC-PR}  & \textbf{F-score}  & \textbf{Precision} & \textbf{Recall}  \\ \hline
LSTM & 0.779 & 0.775 & 0.662 & \textbf{0.751} & 0.593 \\ 
LSTM+$\ell_a$ & 0.785 & 0.786 & 0.688 & 0.704 & 0.684 \\ 
LSTM+$\ell_o$ & 0.804 & 0.821 & 0.702 & 0.673 & 0.737 \\
LSTM+$\ell_a$+$\ell_o$ & 0.817 & \textbf{0.834} & 0.721 & 0.709 & 0.737 \\ 
LSTM+$\ell_a$+$\ell_o$+DA & \textbf{0.818} & 0.823 & \textbf{0.743} & 0.697 & \textbf{0.800} \\ \hline\hline
\end{tabular}
}
\label{tab:ablation}
\end{table}


\section{Conclusions and Future Work}

In this work we propose a novel multi-task learning framework and its instantiation MTNet for depression prediction. MTNet effectively leverages the two auxiliary anomaly ranking tasks to improve the depression prediction model's representations and generalizability. Remarkably, our empirical results show that MTNet is able to accurately predict 80\% depression cases 2-4 years before the depression actually occurs in children. The improved generalizability of MTNet is supported by the sample efficiency experiment, in which MTNet requires significantly less labeled depression samples to perform comparably well to, or substantially better than, the competing methods. It should be noted the model has around 70\% precision only, indicating 30\% false positive predictions. Thus, caution must be taken when using the the model in practice. In future work, to improve the model's accountability and its collaboration with clinical psychologists, we plan to incorporate an interpretation module into our model to provide insightful explanation for each of its depression prediction result.

\bibliographystyle{splncs04}
\bibliography{refs}

\end{document}